\pdfoutput=1

\documentclass[11pt]{article}

\usepackage{emnlp2021}

\usepackage{times}
\usepackage{latexsym}

\usepackage[T1]{fontenc}

\usepackage[utf8]{inputenc}

\usepackage{microtype}

\usepackage{graphicx}
\usepackage{url}
\usepackage{amsmath}
\DeclareMathOperator*{\argmax}{arg\,max}
\usepackage{booktabs}
\usepackage{multirow}
\usepackage{enumitem}
\usepackage{xcolor}

\title{{D}imensional {E}motion {D}etection from {C}ategorical {E}motion}

\author{
 {\bf Sungjoon Park}\textsuperscript{1}
{\bf Jiseon Kim}\textsuperscript{1}
{\bf Seonghyeon Ye}\textsuperscript{1}
{\bf Jaeyeol Jeon}\textsuperscript{2}
{\bf Hee Young Park}\textsuperscript{3}
{\bf Alice Oh}\textsuperscript{1}\\
\textsuperscript{1} School of Computing, KAIST, Republic of Korea \\
\textsuperscript{2} Upstage AI Research, Upstage, Republic of Korea \\
\textsuperscript{3} Department of Psychology, Seoul National University, Republic of Korea \\
{\tt \{sungjoon.park, jiseon\_kim, vano1205\}@kaist.ac.kr},\\
{\tt jaeyeol.jeon@upstage.ai}, ~{\tt heeyoungpark@snu.ac.kr},\\
{\tt alice.oh@kaist.edu}}

\begin{document}
\maketitle
\begin{abstract}

We present a model to predict fine-grained emotions along the continuous dimensions of valence, arousal, and dominance (VAD) with a corpus with categorical emotion annotations. Our model is trained by minimizing the EMD (Earth Mover's Distance) loss between the predicted VAD score distribution and the categorical emotion distributions sorted along VAD, and it can simultaneously classify the emotion categories and predict the VAD scores for a given sentence. We use pre-trained RoBERTa-Large and fine-tune on three different corpora with categorical labels and evaluate on EmoBank corpus with VAD scores. We show that our approach reaches comparable performance to that of the state-of-the-art classifiers in categorical emotion classification and shows significant positive correlations with the ground truth VAD scores. Also, further training with supervision of VAD labels leads to improved performance especially when dataset is small. We also present examples of predictions of appropriate emotion words that are not part of the original annotations.
\end{abstract}

\section{Introduction}
\begin{figure*}[!ht]
\centering
\includegraphics[width=1.0\textwidth]{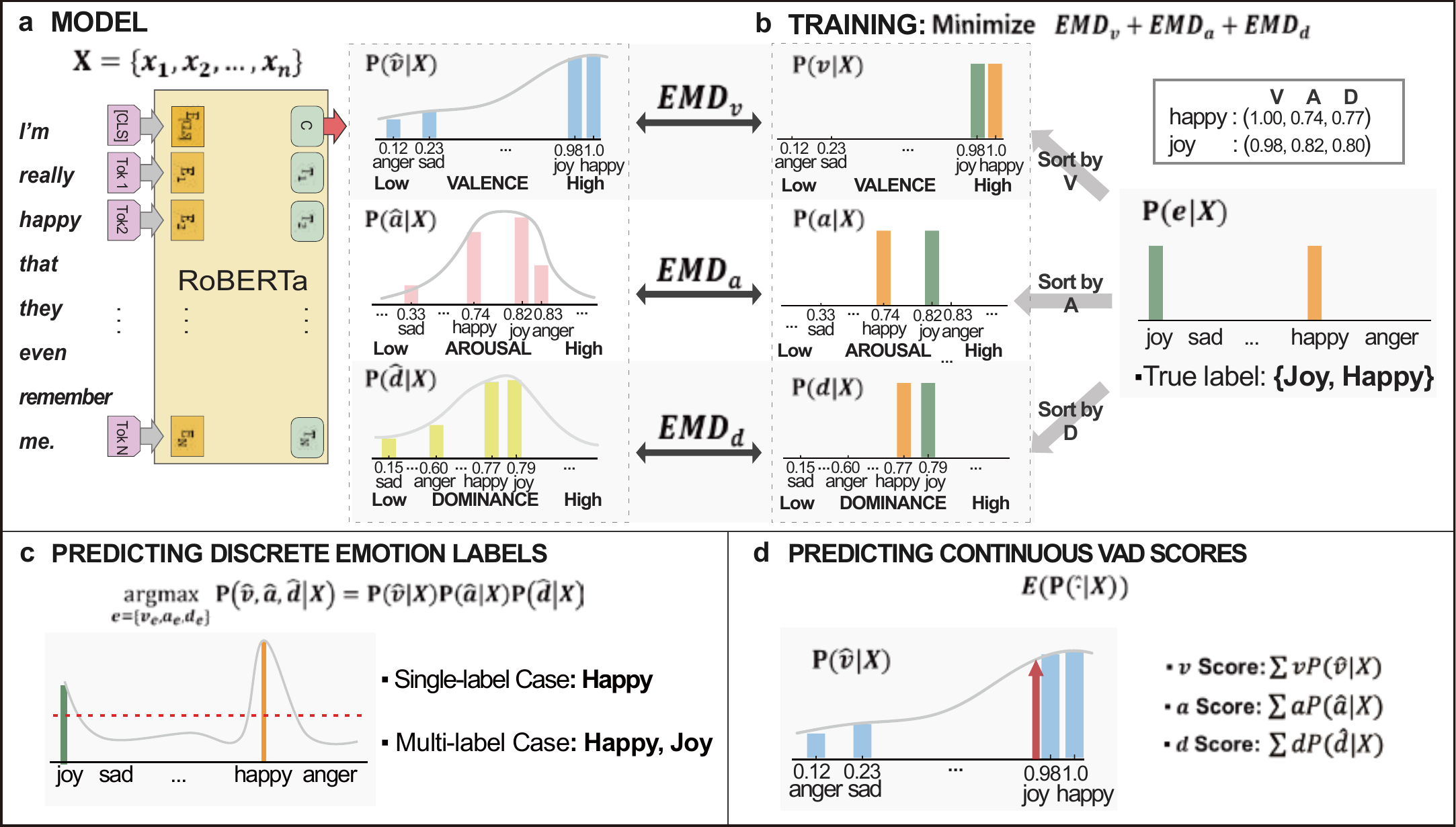}
\caption{Overview of our approach. (a) Our model predicts VAD distributions of input sentence through supervised training with categorical emotion labels. (b) Categorical labels are sorted in terms of VAD scores, to be served as (sparse) label VAD distributions during training. After training, (c) categorical emotion class is predicted by picking one having maximum probability of the product of the three distributions. (d) Continuous VAD scores are predicted by computing expectation of each distribution.}
\label{fig:model}
\vspace{-4mm}
\end{figure*}
 
In psychology literature, \textit{basic} emotions are categorized as \textit{happy, sad, angry} and so on \cite{ekman1992argument, plutchik2001nature}, however, we can feel and express more subtle and complex emotions beyond them. They can be systematically represented with the Valence-Arousal-Dominance (VAD) model which maps emotional states to 3-dimensional continuous VAD space. This space allows various emotions to be projected into the space with measurable distances from one another \cite{russell1977evidence}, covering a wider range of subtle emotions compared to the categorical models with a finite set of basic emotions. 
Capturing such fine-grained emotions with dimensional VAD models could benefit clinical natural language processing (NLP) \cite{Desmet:2013:EDS:2506578.2506869, sahana2015automatic}, emotion regulation such as psychotherapy \cite{doi:10.1177/1754073917742706}. For example, analyzing the client’s utterance and acknowledging the negative emotion as `neglected' rather than `sad', which is known as `affect labeling', would reduce negative physiological, behavioral, and psychological responses resulting from that emotional state.

Thus developing a dimensional emotion detection model would be very useful, but one problem is a lack of required annotated resources. There is a relatively small sentence-level corpus with full VAD annotations \cite{buechel2017emobank}, and a corpus annotated with V and A dimensions \cite{preoctiuc2016modelling, yu2016building}, and only with V \cite{lykousas2019sharing}. We could build additional resources by labeling VAD scores by Best-Worst Scaling \cite{kiritchenko2017best}. Instead, we approach this problem with a novel and more efficient method to predict VAD scores from existing corpora annotated with categorical emotions \cite{scherer1994isear, alm2005tales, aman2007blogs, mohammad2012emotional, sintsova2013olymplex, li2017dailydialog, schuff2017ssec, shahraki2017cbet, SemEval2018Task1}.

In this paper, we propose a framework to learn the VAD scores from sentences with categorical emotion labels by leveraging the VAD scores of the label words obtained from the NRC-VAD lexicon \cite{mohammad2018obtaining}. We demonstrate our approach by fine-tuning a pre-trained language model RoBERTa \cite{liu2019roberta}. Our model learns conditional VAD distributions through supervision of categorical labels and uses them to compute VAD scores as well as to predict the emotion labels for a given sentence. Our contributions are as follows.
\begin{itemize}[noitemsep,nolistsep,leftmargin=*]
    \item We propose a framework which enables learning to predict VAD scores as well as categorical emotions from a sentence only with categorical emotion labels.
    \item Our model shows significant positive correlations to corresponding ground truth VAD scores.
    \item Our model outperforms state-of-the-art dimensional emotion detection models by fine-tuning with supervision of VAD scores when the training dataset size is limited.
\end{itemize}

\section{Approach}
\noindent \textbf{Overview.} We predict VAD scores for a given text from a model trained on a dataset with categorical emotion annotations. The key idea is to train VAD prediction model by using categorical emotion labels. It is possible because we find that those categorical labels can be mapped to word-level VAD scores by using NRC-VAD lexicon \cite{mohammad2018obtaining}. Thus we conceptualize categorical emotion as a \textit{point} in the VAD space. Then we sort the labels by each VAD dimension to obtain (sparse) ground truth conditional VAD distributions (Fig. \ref{fig:model}a, \ref{fig:model}b). Then we train a model to predict the VAD distributions, rather than an emotion category, by minimizing the distance between the predicted and the ground truth distributions. This allows the model to predict the VAD scores (expectations of predicted distributions, Fig. \ref{fig:model}d) and pick an emotion label within a given set of categorical labels (argmax of emotion labels, Fig. \ref{fig:model}c).

\noindent \textbf{Model Architecture (Fig \ref{fig:model}a)}. Formally, an emotion detection model is $P(e|X)$ where $e$ is an emotion drawn from a set of pre-defined categorical emotions $e \in E = \{joy, anger, sadness, ... \}$, and $X=\{x_1, x_2, ..., x_n\}$ is a sequence of symbols $x_i$ representing the input text. Usually $e$ is a one-hot vector in emotion classification. 

Unlike classification models directly learning $P(e|X)$, we learn each distribution of V, A, and D from a pair of input text $X$ and  categorical labels. To this end, we map the categorical emotion labels to the three-dimensional VAD space, $e = (v, a, d)$, using the NRC-VAD Lexicon. Each $v, a$ and, $d$ ranges from 0 to 1. For example, an emotion label "joy" is mapped to (0.980, 0.824, 0.794) and "sad" to (0.225, 0.333, 0.149) \cite{mohammad2018obtaining}. Using $e$s, our model predicts the following distribution:
\begin{equation}
\small{
 P(e|X) = P(v,a,d|X)
}
\label{eq.1}
\end{equation}
Furthermore, since the VAD dimensions are nearly independent \cite{russell1977evidence}, we simply assume mutual independence:
\begin{equation}
\small{
 P(v,a,d|X) \\
=P(v|X)P(a|X)P(d|X).
}
\label{eq.2}
\end{equation}
For each decomposed conditional distribution, we can use any type of trainable function with sufficient complexity to capture the linguistic patterns from the given input. As a demonstration, we use pre-trained bidirectional language model RoBERTa \cite{liu2019roberta} which shows high performances in natural language understanding tasks if fine-tuned over task-specific datasets. We stack a softmax or sigmoid activation layer over the hidden state corresponding to [CLS] token in the model for each conditional distribution.

\noindent \textbf{Model Training (Fig \ref{fig:model}b).} To train our model, we need to obtain target conditionals for each $P(v|X), P(a|X), P(d|X)$ from categorical emotion labels. We simply sort categorical emotions in $E$ by V, A, D scores respectively, based on the mapped VAD coordinates. For example, if we have four emotions in the categorical labels $E=\{joy, sad, happy, anger\}$ and they have corresponding valence (V) scores (0.980, 0.225, 1.000, 0.167) in NRC-VAD \cite{mohammad2018obtaining}, then we sort the labels in the order (anger, sad, joy, happy) and the corresponding one-hot labels to obtain the target conditional $P(v|X)$. In other words, by rearranging the label positions in ascending order of valence scores, sorted one-hot labels can be treated as a \textit{proxy} of target conditionals. Similarly, we sort the labels for the A, D dimensions to obtain the other conditionals. They will be sparse because we only have $|E|$ points for each dimension.

Next, we minimize the distances between the true and predicted $P(\cdot|X)$s. Since we sorted the labels, there is ordering among the classes. This should be taken into account during optimization, so we minimize the squared Earth Mover’s Distance (EMD) loss \cite{hou2017squared} between them to consider the order of labels as follows:
\begin{equation}
\small{
EMD(\mathbf{p}, \hat{\mathbf{p}})=\sum_{c=1}^{C}(CDF_i(\mathbf{p_c}) - CDF_i(\mathbf{\hat{p_c}}) )^{2} 
}
\label{EMD_loss_single}
\end{equation}
where $\mathbf{p}$ is the true conditional, $\mathbf{\hat{p}}$ is the predicted conditional and $c$ is class index. Formally, EMD loss is the squared difference between the cumulative distribution function (CDF) $\mathbf{p}$ and the corresponding $\mathbf{\hat{p}}$. The loss penalizes the mispredictions according to a distance matrix that quantifies the dissimilarities between classes. For instance, if a ground truth is ‘happy’, the loss give more penalty to a prediction ‘sad’ compared to ‘joy’ because ‘sad’ is way more far from ‘happy’ than ‘joy’ on the V dimension. Simple cross-entropy loss cannot reflect this distance between classes.

Note that Eq. \ref{EMD_loss_single} has an assumption that the probability mass of $\mathbf{p}$ and $\mathbf{\hat{p}}$ should be the same. In the single label case, i.e., if the categorical label can appear only once for each text, it could be easilty satisfied when using softmax for $\mathbf{\hat{p}}$. However, in multi-label, this assumption is violated because generally sigmoid is used to represent positive probabilities for each class independently. Thus we slightly change Eq. \ref{EMD_loss_single} to satisfy the assumption, defining interclass EMD loss:
\begin{equation}
\scriptsize{
EMD_{inter}
(\mathbf{p},\hat{\mathbf{p}})
    =\sum_{c=1}^{C} (v_c-v_{c-1})
    ( CDF(\mathbf{\left \langle p_c \right \rangle}) - 
      CDF(\mathbf{\left \langle \hat{p_c} \right \rangle} ))^{2} 
}
\label{interclass_EMD_loss}
\end{equation}
where $\left \langle \mathbf{p_c} \right \rangle$ and $\left \langle \mathbf{\hat{p_c}} \right \rangle$ are corresponding probabilities for class $c$ in normalized $\mathbf{p}$ and $\mathbf{\hat{p}}$. In addition, as shown in Fig. \ref{fig:model}d, the distances between classes are usually not the same, so we give larger weights if they are far from each other through $(v_c-v_{c-1})$. $v_c$ is one of the corresponding V, A, D values for class $c$, and $v_c=0$ if $c=0$. We also introduce intraclass EMD loss:
\begin{equation}
\small{
EMD_{intra}
    (\mathbf{p_c}, \hat{\mathbf{p_c}})
    =\sum_{i=1}^{2}
    ( CDF(\mathbf{p_{ci}}) - 
      CDF(\mathbf{\hat{p_{ci}}} ))^{2} 
}
\label{intraclass_EMD_loss}
\end{equation}
where we assume $\mathbf{p_{c}}$ could be divided into two classes, $[p_c, 1-p_c]$, which represent the probability of belonging to class $c$ : ($p_c$) and not belonging to class $c$ : ($1-p_c$).
\noindent Finally we sum two EMD losses for multi-labeled case as follows:
\begin{equation}
\small{
EMD(\mathbf{p}, \hat{\mathbf{p}})=EMD_{inter} + \frac{1}{C}\sum_{c=1}^{C}EMD_{intra}
}
\label{EMD_loss_multi}
\end{equation}

Finally, we minimize the sum of three squared EMD losses between target and predicted distributions for each of VAD dimensions:
\begin{equation}
\small{
l = EMD(\mathbf{v}, \hat{\mathbf{v}}) + EMD(\mathbf{a}, \hat{\mathbf{a}}) + EMD(\mathbf{d}, \hat{\mathbf{d}})
}
\label{total_loss}
\end{equation}
where $\mathbf{v}$, $\mathbf{a}$, $\mathbf{d}$ denote target and $\mathbf{\hat{v}}$, $\mathbf{\hat{a}}$, $\mathbf{\hat{d}}$ predicted conditional distributions.

\noindent \textbf{Predicting Continuous VAD Scores (Fig. \ref{fig:model}d).} We can further compute the expectations of each predicted conditional distributions of V, A, D dimension to predict the continuous VAD scores.
\begin{equation}
\scriptsize{
    \begin{split}
    v_X = E(\hat v) = \sum_{i=1}^{C} v_i P(\hat v_i|X),~~~
    a_X = E(\hat a) = \sum_{i=1}^{C} a_i P(\hat a_i|X), \\
    d_X = E(\hat d) = \sum_{i=1}^{C} d_i P(\hat d_i|X)~~~~~~~~~~~~~~~~~~~~~~~~~~~~~~
    \end{split}
    }
\label{vad_expectation}
\end{equation}
Once again, we use the VAD scores in \cite{mohammad2018obtaining} for each dimension when computing the expectations. This allows us to predict continuous VAD scores from the model which is trained over categorical emotion annotations. 

\noindent \textbf{Predicting Categorical Emotion Labels (Fig. \ref{fig:model}c).} We can further recover categorical emotions from the predicted distributions. We pick one emotion label from a given set $E$ as in the conventional emotion classifiers. By computing the product of predicted $p(v|X)$, $p(a|X)$, $p(d|X)$, we obtain predicted $p(v,a,d|X)$ assuming conditional independence. Then we pick an emotion label $e \in E$ as follows:
\begin{equation}
\small{
    \argmax_{\{v,a,d\}=e \in E}P(v,a,d|X)
    }
\label{eq:argmax}
\end{equation}
Since we only have $|E|$ given emotion labels, we compare the joint probabilities of $(v,a,d)=e \in E$ and pick one emotion label having the maximum probability among labels (single-label case, Eq. \ref{eq:argmax}), or multiple labels with probability over a certain threshold (multi-label case). The threshold is a hyperparameter of the model, set to $0.5^{1/3}$, a geometric mean of the three distributions.

\section{Experiments}
We mainly focus on demonstrating our approach can effectively predict continuous emotional dimensions only with categorical emotions throughout experiments. 

\subsection{Dataset} 
We use four existing datasets consisting of text and corresponding emotion annotations. Three of them have categorical emotion labels, and the last is annotated with VAD scores.

\noindent \textbf{SemEval 2018 E-c (SemEval).} A multi-labeled categorical emotion corpus contains 10,983 tweets and corresponding labels for presence or absence of 11 emotions \cite{SemEval2018Task1}. We abbreviate this hereafter as SemEval. We use pre-splits of train, valid, test set of the dataset.

\noindent \textbf{ISEAR.} A single-labeled categorical emotion annotated corpus contains 7,666 sentences. A label can have only one emotion among 7 categorical emotions \cite{scherer1994isear}. We split the dataset in a stratified fashion in terms of the labels. The train, valid, test set is split by the ratios (0.7:0.15:0.15).

\noindent \textbf{GoEmotions.} A multi-labeled categorical dataset consisted of of 58,009 reddit comments with 28 emotion labels including \emph{neutral} class \cite{demszky2020goemotions}. However, the original dataset with 28 emotion labels has large disparity in terms of emotion frequencies (\emph{admiration} is 30 times more frequent that \emph{grief}). To reduce the side-effects from this property, we choose the `Ekman' option of the dataset provided by the authors which consists of 7 emotion labels including \emph{neutral} class. We use pre-splits of train, valid, test set of the dataset.

\noindent \textbf{EmoBank.} Sentences paired with continuous VAD scores as labels. This corpus contains 10,062 sentences collected across 6 domains and 2 perspectives. Each sentence has three scores representing VAD in the range of 1 to 5. Unless otherwise noted, we use the weighted average of VAD scores as ground truth scores, which is recommended by EmoBank authors. We use pre-splits of train, valid, test set of the dataset \cite{buechel2017emobank}.

\subsection{Dimensional Emotion (VAD) Prediction}
We investigate VAD score prediction performance of our approach and compare them to the state-of-the-art models. Since training objectives of models vary, we use Pearson's correlation coefficient as the evaluation metric.

\subsubsection{Zero-shot VAD Prediction}
\noindent \textbf{Our Models.} We compute the VAD score predictions using Eq. \ref{vad_expectation} with our model trained on three datasets with categorical emotion annotations (SemEval, ISEAR, GoEmotions). We call these results as \textit{zero-shot} prediction performances because they are not trained over the EmoBank train-set, only using the EmoBank test-set for evaluation. This could be interpreted as how much a model can \textit{generalize} the categorical emotions into the continuous VAD space using only $|E|$ fixed points in the space. These are denoted as \textbf{(Ours, $\mathbf{d}$)} where $d \in \{$SemEval, ISEAR, GE$\}$ in Table.~\ref{table:vadresults}. We highlight these results to evaluate our main idea.

\subsubsection{VAD Prediction with Supervision}
We continuously train the our zero-shot models with the train-set of the EmoBank, and compare their performance with other methods which relies on the direct supervision from them. This allow us to compare the zero-shot prediction performances against them, and how much the zero-shot prediction model could be improved if VAD annotations are available. We also compare data scarce scenarios, only using a part of Emobank training-set.

\noindent \textbf{Our Models.} We fine-tune our zero-shot models once again on the Emobank train-set. In the first stage, we train zero-shot models by combining the EMD loss with MLM loss to prevent catastrophic forgetting \cite{chrono2019transfer}. In the second stage, we add another linear layer and ReLU activations on top of the model for each VAD dimensions. All of the parameters are fine-tuned by minimizing the mean squared error loss (MSE). During fine-tuning, parameters are freezed for 5 epochs except the added linear layer and then all parameters are unfreezed. Through this model, we investigate the effectiveness of our approach as a parameter initialization strategy of VAD regression model where the VAD annotations are available. These models are denoted as \textbf{(Ours, EB$\mathbf{\leftarrow d}$)} where $d \in \{$SemEval, ISEAR, GE$\}$ in Table. \ref{table:vadresults}.

\noindent \textbf{AAN.} Adversarial Attention Network for dimensional emotion regression which learns to discriminate VAD dimension scores \cite{zhu2019adversarial}. Pearson correlations of predicted and ground truth of VAD scores in EmoBank are reported. Since the scores are reported by 2 perspectives and 6 domains respectively, we use the highest VAD correlations among perspective and domains.

\noindent \textbf{Ensemble.} Multi-task ensemble neural networks which learns to predict VAD scores, sentiment, and their intensity simultaneously \cite{8756111}.

\noindent \textbf{SRV-SLSTM.} Predicting VAD scores through variational autoencoders trained by semi-supervised learning, which shows state-of-the-art performance on the VAD score prediction task \cite{wu2019semi}. The model shows highest performance when using 40\% of labeled Emobank data, so we compare our model's performances to scores of that setting.

\noindent \textbf{RoBERTa-Large (Regression).} We add simple yet effective baseline for fair comparison. We add a linear layer with Relu on top of pre-trained RoBERTa \cite{liu2019roberta} for training on a entire EmoBank training-set. The models are optimized by minimizing the mean squared error loss (MSE).

\subsection{Categorical Emotion Prediction} 
We examine classification performances of our approach and compare them to the state-of-the-art emotion classification models. We use accuracy and macro/micro F1 scores as evaluation metrics.

\noindent \textbf{Our Models.} We fine-tune RoBERTa with our EMD objective and predict the emotion category as shown in Fig~\ref{fig:model}c. For a multi-labeled dataset (SemEval, GoEmotions), we minimize Eq.~\ref{total_loss} with Eq.~\ref{EMD_loss_multi}. For a single-labeled dataset (ISEAR), we fine-tune RoBERTa by minimizing Eq.~\ref{total_loss} with Eq.~\ref{EMD_loss_single} for each VAD dimension. These models are denoted as \textbf{($\mathbf{d}$, M)} where $d \in \{$SemEval, ISEAR, GoEmotions$\}$ and $M \in \{$state-of-the-art, RoBERTa, Ours$\}$ in Table.~\ref{table:classresults}.

\noindent \textbf{MT-CNN.} A convolutional neural network for text classification trained by multi-task learning \cite{zhang2018text}. The model jointly learns classification labels and emotion distributions of a given text. The model reaches state-of-the-art classification accuracy and F1 score on ISEAR.

\noindent \textbf{NTUA-SLP.} A classifier using deep self-attention layers over Bi-LSTM hidden states. The models is pre-trained on general tweets and `SemEval 2017 task 4A', then fine-tuned over all `SemEval 2018 subtasks' \cite{baziotis2018ntua}. The model took first place in multi-labeled emotion classification task on SemEval.

\noindent \textbf{Seq2Emo} A sequence-to-sequence model for multi-label classification task. \cite{huang2019seq2emo}. The model additionally leverages correlations between emotion labels during classification.

\noindent \textbf{RoBERTa-Large (Classification).} As a simple baseline, we add a linear layer with sigmoid activation on RoBERTa \cite{liu2019roberta} for training on a multi-labeled dataset (SemEval, GoEmotions) or softmax activation for single-labeled dataset (ISEAR). These models are optimized by minimizing the cross-entropy loss.

\subsection{Experimental Details} 
In all experiments, we use PyTorch version of RoBERTa-Large from Huggingface Transformers \cite{Wolf2019HuggingFacesTS}. We set the learning rate to 3e-5, batch size to 32. Fine-tuning parameters is stopped when the validation loss and and evaluation metrics are converged. We use 1 RTX 6000 GPU for optimization. More details are in Appendix. We release our implementation in GitHub.~\footnote{\scriptsize{\url{https://github.com/SungjoonPark/EmotionDetection}}}

\begin{table}[!t]
\small
\begin{tabular}{@{}l|lrrr@{}}
\toprule
Dataset & \multicolumn{4}{c}{\begin{tabular}[c|]{@{}c@{}}\textbf{EmoBank} \\ \tiny{\cite{buechel2017emobank}}\end{tabular}}  \\
Task & \multicolumn{4}{c}{Regression} \\ \midrule
Model & \multicolumn{1}{c|}{Scheme} & \multicolumn{1}{c}{V (r)} & \multicolumn{1}{c}{A (r)} & \multicolumn{1}{c}{D (r)} \\
\midrule
Ours, SemEval & \multicolumn{1}{l|}{\tiny{Zero-shot}} & \textbf{.715} & \textbf{.319} & .308 \\
Ours, ISEAR & \multicolumn{1}{l|}{\tiny{Zero-shot}}& .611 & .083 & .242\\ 
Ours, GE & \multicolumn{1}{l|}{\tiny{Zero-shot}}& .630 & .277 & \textbf{.311} \\ \midrule
AAN 
& \multicolumn{1}{l|}{\tiny{Supervised}} & .424 & .352 & .265 \\
Ensemble 
& \multicolumn{1}{l|}{\tiny{Supervised}} & .635 & .375 & .277\\
SRV-SLSTM 
& \multicolumn{1}{l|}{\tiny{Semi-super.}} & .620 & .508 & .333\\ \midrule

RoBERTa-Large & \multicolumn{1}{l|}{\tiny{Supervised}} & .829 & .569 & .513 \\
Ours, EB$\leftarrow$SemEval & \multicolumn{1}{l|}{\tiny{Supervised}} & \textbf{.838} & .570 & .518 \\
Ours, EB$\leftarrow$ISEAR & \multicolumn{1}{l|}{\tiny{Supervised}} & .836 & .568 & \textbf{.536}\\
Ours, EB$\leftarrow$GE & \multicolumn{1}{l|}{\tiny{Supervised}} & .835 & \textbf{.573} & .529\\
\bottomrule
\end{tabular}
\caption{Performance of VAD score prediction. With fine-tuning pre-trained RoBERTa-Large, we show significant positive correlations with VAD scores using only the categorical emotion annotations. If those models are continuously fine-tuned on EmoBank, it outperforms all SOTA VAD regression models. Validation set results are in Appendix.}
\label{table:vadresults}
\vspace{-4mm}
\end{table}

\section{Results}
\noindent \textbf{Zero-Shot VAD Prediction.}
The results are shown in Table \ref{table:vadresults}.
When our model is trained on SemEval and tested on Emobank, the predicted VAD scores show significant positive Pearson's correlation coefficients with target VAD scores in EmoBank. 
The correlation in valence (V) shows the highest score among the dimensions (\textit{r}=.715, \textit{p}$<$.001), followed by arousal (A) (\textit{r}=.319, \textit{p}$<$.001), and dominance (D) (\textit{r}=.308, \textit{p}$<$.001). 
For our model trained on ISEAR dataset, the scores also show significant positive Pearson's $r$. The correlation in V dimension is highest (\textit{r}=.611, \textit{p}$<$.001), followed by D (\textit{r}=.242, \textit{p}$<$.001), and A (\textit{r}=.083, \textit{p}$<$.001). For GoEmotions dataset, the highest correlation is also V dimension (\textit{r}=.630, \textit{p}$<$.001), followed by dominance (D) (\textit{r}=.311, \textit{p}$<$.001), and arousal (A) (\textit{r}=.277, \textit{p}$<$.001). We observe prediction performances of VAD scores from text usually are the best in V dimension and A, D follows. These tendencies are observed in our experiments as well as in other baselines (AAN, Ensemble, SRV-SLSTM).

The average of correlations between dataset is in the order of SemEval (.448), GoEmotions (.406), and ISEAR (.312) in descending order. The main reason SemEval has best performance is because emotion labels in SemEval have more information than that of ISEAR or GoEmotions. First, SemEval has 11 categorical emotion annotations whereas ISEAR and GoEmotions have 7 labels each.
More labels lead to less sparse VAD target distributions, thus our model can distinguish the extent of VAD more easily where there are more labels.
Second, SemEval and GoEmotions can have multiple emotion labels for every sentence, but ISEAR has only one label. Apparently, these multiple emotion labels makes the possible range of the expected VAD scores much wider than that of single emotion labels. 
If a sentence always should have a single label, then the predicted VAD distribution must sum up to one. 
Having multiple labels enables the distributions to sum to a larger number, which leads to a wider range of the expected values that help the model distinguish the degree of VAD dimensions for a given sentence. 

Note that we observe the correlation in A dimension of ISEAR is low. 
We see that the standard deviation of arousal scores of ISEAR labels {`anger', `disgust', `fear', `sadness', `shame', `joy', 'guilt'} is lower (.191) than other dimensions, (V: .313, D: .235) and actually it becomes much lower when only one label 'sadness', is removed, dropping to (.105). 
This makes model difficult to differentiate labels in terms of the degree of arousal, leading to lower correlation for the A dimension.

\begin{figure}[!t]
\centering
\includegraphics[width=0.4\textwidth]{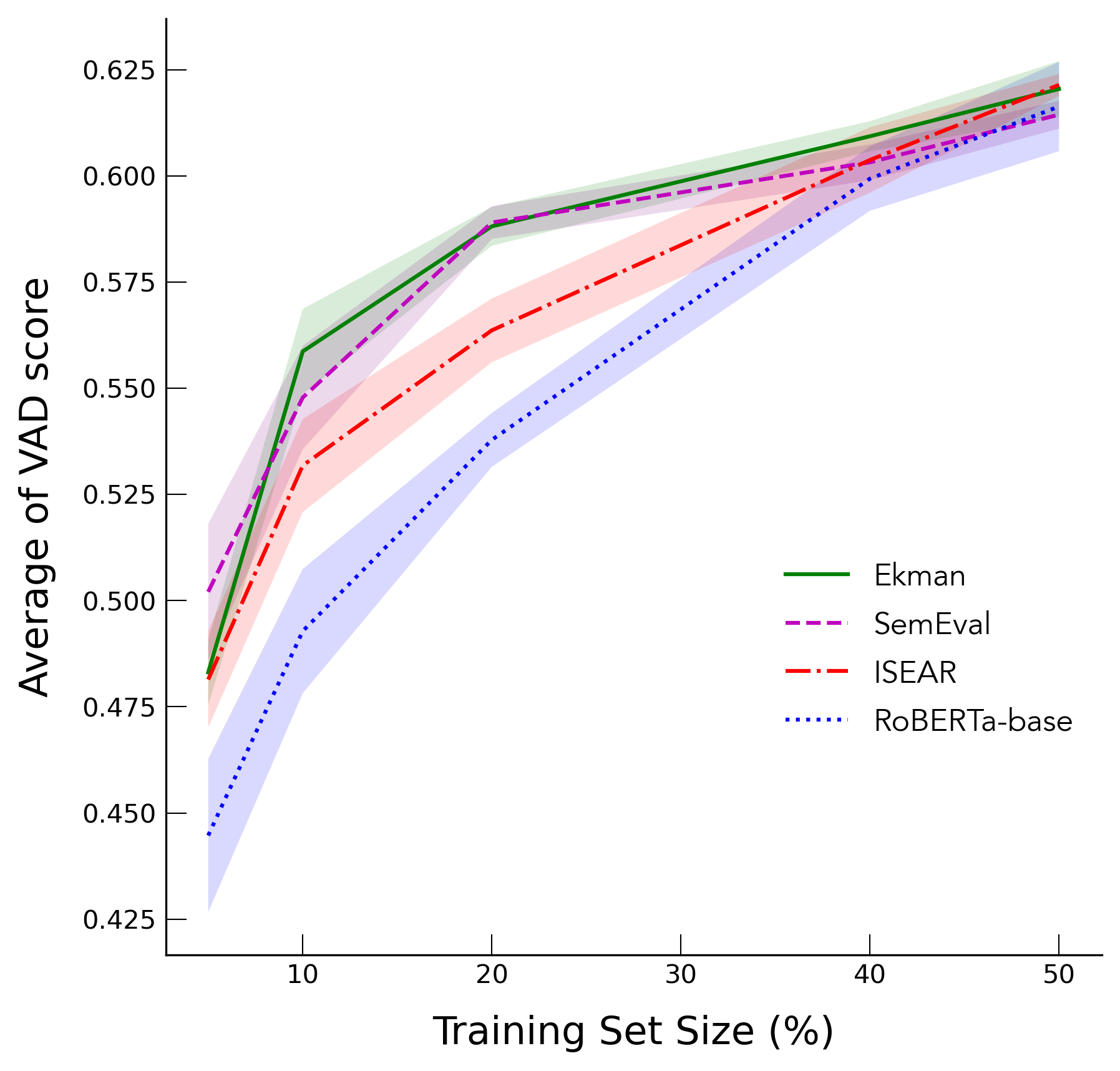}
\caption{Average VAD prediction score when using a part of EmoBank training data. Our model performs better compared to RoBERTa when less data is available. The error bars mark the region within 1 standard deviation and the lines indicate the average of five runs.}
\label{fig:fewshot}
\vspace{-4mm}
\end{figure}

\noindent \textbf{VAD prediction with Supervision.}
Three comparison models (AAN, Ensemble, SRV-SLTSTM) in Table \ref{table:vadresults} are trained by supervision of VAD scores. Among the comparison models, Ensemble shows the highest correlation on V dimension (.635), SRV-SLSTM reaches to the highest correlation on A (.508) and D (.333) dimensions. We emphasize that our model trained on SemEval shows even better correlation in the V dimension (.715) without any supervision of VAD scores.
Correlation for A (.319) is  next which is slightly lower than AAN and Ensemble, and correlation for D (.308) is comparable to SRV-SLTSTM.

Furthermore, we observe that if we continue training our zero-shot models with supervision of VAD labels, our model outperforms all of the state-of-the-art models with a large margin.
For model trained on SemEval, the VAD fine-tuned model shows a significant correlation in V (\textit{r}=.838, \textit{p}$<$.001), A (\textit{r}=.570, \textit{p}$<$.001) and D (\textit{r}=.518, \textit{p}$<$.001) dimensions.
For ISEAR, the fine-tuned model shows correlation of V (\textit{r}=.836, \textit{p}$<$.001), A (\textit{r}=.568, \textit{p}$<$.001) and D (\textit{r}=.536, \textit{p}$<$.001) dimensions. 
For GoEmotions, the fine-tuned model shows correlation of V (\textit{r}=.835, \textit{p}$<$.001), A (\textit{r}=.573, \textit{p}$<$.001) and D (\textit{r}=.529, \textit{p}$<$.001) dimensions. 
The average of supervised result between dataset is in the order of ISEAR (.647), GoEmotions (.646) and SemEval (.642) in descending order. For model trained from ISEAR, these are (+.215, +.065, +.196) improvement of the correlation from the state-of-the-art models with supervision for VAD dimensions.

In fact, the performance of our approach are comparable to that of RoBERTa-large (Regression) and it shows  correlations in V (\textit{r}=.829, \textit{p}$<$.001), A (\textit{r}=.569, \textit{p}$<$.001) and D (\textit{r}=.513, \textit{p}$<$.001) dimensions. We see that this is because the size of Emobank training set is sufficiently large, so we further conduct experiment assuming the training data is small. Figure \ref{fig:fewshot} shows results on such settings, using only \{5, 10, 20, 30, 40, 50\}\% of the training data. For all models initialized to our fine-tuned model on SemEval, ISEAR, Ekman, our method shows better performance compared to that of RoBERTa-large (Regression) when using of training data is smaller.

\begin{table}[!t]
\small
\begin{tabular}{@{}l|rrr@{}}
\toprule
Dataset (Model) & \multicolumn{1}{c}{\begin{tabular}[c]{@{}c@{}}Macro \\ F1\end{tabular}} & \multicolumn{1}{c}{\begin{tabular}[c]{@{}c@{}}Micro \\ F1\end{tabular}} & \multicolumn{1}{c}{Acc.} \\
\midrule
ISEAR (MT-CNN) & - & .668 & - \\
ISEAR (RoBERTa) & .754 & .755 & - \\ 
ISEAR (Ours) & .752 & .753 & -\\ \midrule
SemEval (NTUA-SLP) & .528 & .701 & .588 \\
SemEval (Seq2Emo) & - & .709 & .592 \\
SemEval (RoBERTa-Large) & .574 & .725 & .607 \\
SemEval (Ours) & .566 & .725 & .607 \\ \midrule
GoEmotions \scriptsize{\cite{demszky2020goemotions}} & .640 & - & -\\
GoEmotions (RoBERTa) & .618 & .691 & .659 \\ 
GoEmotions (Ours) & .611 & .686 & .657 \\ 
\bottomrule
\end{tabular}
\caption{Performance of categorical emotion classification. With fine-tuning pre-trained RoBERTa, we show comparable performance to SOTA models in classification. Validation set results are in Appendix.}
\label{table:classresults}
\vspace{-4mm}
\end{table}

\noindent \textbf{Categorical Emotion Prediction.}
Next, classification performances of our model and that of comparison models are reported in Table. \ref{table:classresults}.
Note that our model outperforms all baseline models for emotion classification except RoBERTa-Large, which is comparable to our model.

\section{Ablation Study}
\begin{table}[!t]
\small
\begin{tabular}{@{}l|rrr|r@{}}
\toprule
Model & V (r) & A (r) & D (r) & Avg. \\ \midrule
\textbf{Zero-Shot} & & & \\
~~1. RoBERTa \scriptsize{(CE, SE)} & .685 & .315 & .278 & .426 \\
~~2. RoBERTa \scriptsize{(Ours, SE)} & .715 & .319 & .308 & .448 \\
\midrule
\textbf{Supervised} & & & \\
~~3. RoBERTa \scriptsize{(Random, EB)} & .381 & .386 & .253 & .340 \\ 
~~4. BERT \scriptsize{(Pretrained, EB)} & .794 & .537 & .514 & .615 \\ 
~~5. RoBERTa \scriptsize{(Pretrained, EB)} & .829 &	.569 & .513 & .637 \\
~~6. Ours, SE-EB \scriptsize{(RoBERTa)} & .838 & .570 & .518 & .642 \\

\bottomrule
\end{tabular}
\caption{Ablation study results. Given that the model architecture is the same (RoBERTa-Large), the architecture and its pre-trained knowledge are effective for VAD regression. Overall, initialization with our model trained on categorical emotions (SE, SemEval) and then fine-tuning on VAD (EB, EmoBank) helps improve performance. Validation set results are in Appendix.}
\label{table:transfer_results}
\vspace{-5mm}
\end{table}

\begin{table*}[h]
\small
\begin{tabular}{l|l|l}
\toprule
\textbf{Tweet} & \textbf{Categorical Label} & \textbf{Nearest Neighbors from VAD scores} \\ \midrule
\begin{tabular}[c]{@{}l@{}}Gooood morning it is such a \#blessing to see another day \\ all that Read this I hope have a great morning\end{tabular} & joy, optimism & \begin{tabular}[c]{@{}l@{}}reaffirm, shimmer, \\ brighten, affections, mythological\end{tabular} \\ \midrule
\begin{tabular}[c]{@{}l@{}}Not only was and responsible for the \\ unnecessary outrage of this movie, \\ but made the director look bad\end{tabular} & anger, disgust & \begin{tabular}[c]{@{}l@{}}refusal, liar, falsified, \\ disrespect, unsavory\end{tabular} \\ \midrule
\begin{tabular}[c]{@{}l@{}}Mentally suffered \#iwanttodie \#worthless \\ \#lifewithoutcolor \#pain \#suicidal\end{tabular} & \begin{tabular}[c]{@{}l@{}}disgust, pessimism, \\ sadness\end{tabular} & \begin{tabular}[c]{@{}l@{}}orphaned, wasting, decomposed, \\ hopelessness, dead\end{tabular} \\ \bottomrule
\end{tabular}
\caption{Qualitative examples of predictions from our model trained on SemEval. Examples Tweets are from test set of SemEval. We present predicted categorical emotion labels, and corresponding top 5 nearest neighbor words in NRC-VAD-Lexicons with respect to the model predictions of VAD scores.}
\label{qualitativeExamples}
\vspace{-5mm}
\end{table*}

We further conduct ablation study to investigate our model's VAD prediction performances.
Since we use pre-trained RoBERTa and fine-tune them with different datasets, the effect of model architecture, pre-training and fine-tuning should be decomposed to understand the source of improvements. We show the result for SemEval dataset because it gave the best performance for zero-shot score prediction. Validation set results are shown in Appendix.

In Table \ref{table:transfer_results}, we present six models for ablation study. 
Model 1 is RoBERTa trained on SemEval with our framework except EMD loss replaced with cross-entropy which does not consider the order of classes in terms of VAD. Compared to Model 2, RoBERTa trained on SemEval with EMD loss, our model shows better correlations in overall. (+.022)

Model 3 is fine-tuned on EmoBank without pre-trained weights of RoBERTa, showing highly underperforming result compared to Model 5, which take advantage of pre-trained weights. Still the performance of Model 3 is comparable to that of AAN \cite{zhu2019adversarial}, it could be highly improved with using pre-trained knowledge obtained from masked language modeling task. (+.302) More Interestingly, Model 4 uses BERT \cite{BERT} pre-trained weights, showing slightly lower performance than Model 5. This indicates using better language models also improves the performance. (+.027) Model 6 shows comparable performance compared to Model 5 when using full 
train-set. 

\section{Qualitative Examples}
In Table \ref{qualitativeExamples}, we show examples predicted from our model trained on SemEval.
The table presents annotated tweets from SemEval test set, corresponding predicted categorical labels, and top 5 nearest neighbor emotional words with respect to the predicted VAD scores.
For these three tweets, our model correctly predicted the categorical emotion labels.
We elaborate how we find the nearest neighbor words from the VAD scores.

Given our model's predicted VAD scores, we find the nearest neighbor words for those scores by using NRC-VAD-Lexicons \cite{mohammad2018obtaining}.
We first rescale our model's predicted VAD scores from 0 to 1 for each VAD dimension since the NRC-VAD lexicons have values from 0 to 1.
To do this, we first predict VAD scores for every sentence in SemEval test set and then we rescale the scores by following: ($x - min(x)) / (max(x) - min(x))$ to make all dimension scores range from 0 to 1.

Next, we find the nearest neighbor words by using the rescaled VAD values. 
Euclidean distances between the values and all words in NRC-VAD-Lexicons are computed, and we pick the top five nearest words among them with the smallest distances.
We present the words in the right column of Table \ref{qualitativeExamples}. Note that these words are extracted from NRC-VAD lexicons so some words are not emotional because it contains frequently used 20,000 English words. However, these words help us understand VAD scores intuitively, and they could be regarded as automatically generated emotion annotations for a given sentence, which are \textit{not} seen during training.


\section{Related Work}
Categorical model of emotion assumes that the categories represented by emotion words compose the building blocks of human emotion. Supporting evidence includes six basic emotions \cite{ekman1992argument}, and findings of universally adaptive emotions \cite{plutchik1980general}. An alternative to understand how people conceptualize emotional feelings is the dimensional model of emotion. \citet{osgood1957measurement} suggested the initial idea of emotion coordinates. \citet{russell1977evidence} further constructed Pleasure or Valence-Arousal-Dominance (PAD, VAD) model, a semantic scale model to rate emotional state, representing an emotional state as a pair of orthogonal coordinates on VAD dimensions. Absolute values of the intercorrelations among the three scales show considerable independence among the scales \cite{russell1977evidence}, categorical emotion states can be represented in 3D (VAD) emotion space. 

Based on emotional dimensions, word-level VAD annotation of English words has been created \cite{bradley1999affective, Warriner2013, mohammad2018obtaining}. Also, there are few sentence-level VA or VAD annotated corpora \cite{buechel2017emobank, preotiuc-pietro-etal-2016-modelling, yu-etal-2016-building}. By using these resources, recent work tried to predict VAD scores from sentences based on variational autoencoders \cite{wu2019semi}, adversarial learning \cite{zhu2019adversarial}, ensemble learning \cite{8756111}. However, sentence-level VAD annotated corpus is scarce, we use more common resource which is sentences annotated with basic categorical emotions for VAD score prediction \cite{scherer1994isear, alm2005tales, aman2007blogs, mohammad2012emotional, sintsova2013olymplex, li2017dailydialog, schuff2017ssec, shahraki2017cbet, SemEval2018Task1, demszky2020goemotions}. These datasets are commonly used for emotion classification, we use them to predict VAD scores from sentences with word-level VAD scores of categorical emotion labels.

Recently, a lot of dataset related to emotion has been released. Especially, there are dataset in healthcare domain \cite{sosea2020canceremo}, relation between emoji and emotion \cite{shoeb2020emoji}, and emotional text from social media \cite{ding2020socialemotion}. All of these are cateogrical annotations which again shows the lack of dimensional annotations thus the need for our model to capture fine-grained emotion detection. Also, our work could be extended to a large domain: it could help better performance of multimodal emotion detection \cite{zhang2020multimodal}, emotion in conversation \cite{ishiwatari2020erc}, and emotion change in a paragraph \cite{brahman2020protagonist}.


There are multiple emotion datasets annotated with various types of label sets. To train model across the various shaped emotion dataset, several existing studies aggregate various format of emotion dataset into a common annotation schema, and show better performance using unified dataset \cite{bostan-klinger-2018-analysis,  belainine-etal-2020-towards}.  However, still the labels are mapped to other predefined emotions and the datasets are limited to categorical labels.
In \cite{buechel-hahn-2018-emotion}, they convert categorical emotions into VAD representation using simple Feed-Forward Neural Networks. They train model with dataset labeled with both emotion categories and VAD. However, in our paper, we convert categorical emotion knowledge to VAD without any labeled pairs.

%


\section{Discussion and Conclusions} 
We propose learning to predict VAD scores from the text with categorical emotion annotations. Our framework predicts VAD score distributions by minimizing the EMD distances between predicted VAD distributions and sorted label distributions as a proxy of target VAD distributions. Even our model assumes VAD emotion space and order between emotions, our model shows significant prediction performances in real-world datasets.

\noindent \textbf{Robustness.} Our framework could be applied to multimodal datasets. If we apply our framework to IEMOCAP \cite{busso2008iemocap}, the zero-shot VAD predictions are significantly correlated with ground truths (V: 0.396, A: 0.241, D: 0.197) as well. However, the performance is rather low since our model does not leverage other modalities such as audio or videos. Once our framework is extended to integrate such information through image/speech encoders, performance would be improved. We use NRC-VAD to estimate distance between emotions because it is constructed very carefully to locate words in VAD space. If we use other word-level VAD resources such as ANEW \cite{redondo2007spanish}, we observe positive results as well (V: 0.682, A: 0.270, D: 0.296).

\noindent \textbf{Ethical Considerations.} A model trained by our approach could be used to understand and regulate one's own emotional states and to save people from suicide. In addition, social bots capable of emotion recognition could help people in various ways. However, a model trained by our approach could be misused to detect or control others' emotional states against their will. It may reveal private information about mental or physical health or private feelings an individual does not wish to share. This concern is even more serious when we consider that machine learning models can be cost-effective and thus used at scale for pervasive monitoring of emotions \cite{PAIreport}. An example of a harmful use of the technology is manipulating the semantic emotive content of user news feeds which can affect the choices of both individuals and groups on the platform to engage and interact \cite{stark_hoey_2020}. From a different perspective, problems might occur from the inaccurate results of the model. Mispredictions of the models could result in harmful outcomes even in systems designed to be helpful, and this is a serious problem in many languages with relatively low resources (i.e., languages other than English and a few others that are extensively studied within NLP), resulting in inequity with respect to the benefits gained by this technology. Basically, resources to train emotion detection models are scarce in most languages, and their quality would degrade if translated to other languages from English since cultural nuances to defining emotions vary. Therefore, one should follow guidelines for the ethical use of emotional AI technologies, which present a checklist for anyone engaged with data about human emotion. \cite{stark_hoey_2020} For example, McStay and Pavliscak’s guidelines \cite{eai_guideline} include a number of salutary suggestions for taking action as a practitioner.

We hope our framework will be helpful in building an annotated sentence-level VAD emotion dataset by providing machine-annotated VAD scores as a start, or use it just as VAD score prediction model. Most of the languages except English would not have such corpus with VAD annotations, so our model will be helpful to build resources using multilingual corpora with categorical emotion labels \cite{ohman2018creating}.

\section*{Acknowledgements}
This research was supported by the Engineering Research Center Program through the National Research Foundation of Korea (NRF) funded by the Korean Government MSIT (NRF2018R1A5A1059921).

\clearpage
\bibliography{anthology,custom}
\bibliographystyle{acl_natbib}

\clearpage
\appendix
\section{Appendix}
\label{sec:appendix}

\begin{table}[h]
\small
\begin{tabular}{@{}l|rrr|r@{}}
\toprule
Model & V (r) & A (r) & D (r) & Avg. \\ \midrule
\textbf{Zero-Shot} & & & & \\
~~1. RoBERTa \scriptsize{(CE, SE)} & .682 &	.310 & .249	& .414 \\
~~2. RoBERTa \scriptsize{(Ours, SE)} & \textbf{.710}& \textbf{.327} & \textbf{.282} & \textbf{.440} \\
\midrule
\textbf{Supervised} & & & & \\
~~3. RoBERTa \scriptsize{(Random, EB)}  & .400 & .411 &	.184 & .332\\
~~4. BERT \scriptsize{(Pretrained, EB)} & .806 & .596 &	.473 & .625\\ 
~~5. RoBERTa \scriptsize{(Pretrained, EB)} & \textbf{.839} & \textbf{.605} & .512 & \textbf{.652}\\
~~6. RoBERTa \scriptsize{(Ours, EB-SE)} & .834& .594 &	\textbf{.517} & .648\\

\bottomrule
\end{tabular}
\caption{Validation Set performance of models in Ablation study.}
\label{table:val_transfer_results}
\vspace{-5mm}
\end{table}

 \begin{table*}[h]
 \scriptsize
 \begin{tabular}{@{}l|lrrr|rrr|rr|rrr@{}}
 \toprule
  Dataset & \multicolumn{4}{c|}{\textbf{EmoBank}} & \multicolumn{3}{c|}{\textbf{SemEval 2018 E-c}} & \multicolumn{2}{c|}{\textbf{ISEAR}}& \multicolumn{3}{c}{\textbf{GoEmotions (GE)}} \\
 Task & \multicolumn{4}{c|}{Regression} & \multicolumn{3}{c|}{\begin{tabular}[c|]{@{}c@{}}Classification \\ ($|E|$=11)\end{tabular}} & \multicolumn{2}{c|}{\begin{tabular}[c]{@{}c@{}}Classification \\ ($|E|$=7)\end{tabular}} &
 \multicolumn{3}{c}{\begin{tabular}[c|]{@{}c@{}}Classification \\ ($|E|$=7)\end{tabular}} \\ \midrule
 Model & \multicolumn{1}{c|}{Scheme} & \multicolumn{1}{c}{V (r)} & \multicolumn{1}{c}{A (r)} & \multicolumn{1}{c|}{D (r)} & \multicolumn{1}{c}{\begin{tabular}[c]{@{}c@{}}Macro \\ F1\end{tabular}} & \multicolumn{1}{c}{\begin{tabular}[c]{@{}c@{}}Micro \\ F1\end{tabular}} & \multicolumn{1}{c|}{Acc.} & \multicolumn{1}{c}{\begin{tabular}[c]{@{}c@{}}Macro \\ F1\end{tabular}} & \multicolumn{1}{c|}{\begin{tabular}[c]{@{}c@{}}Micro \\ F1\end{tabular}} &
 \multicolumn{1}{c}{\begin{tabular}[c]{@{}c@{}}Macro \\ F1\end{tabular}} & \multicolumn{1}{c}{\begin{tabular}[c]{@{}c@{}}Micro \\ F1\end{tabular}} & \multicolumn{1}{c}{Acc.} \\
 \midrule
RoBERTa-Large \tiny{(Classification)} & \multicolumn{1}{c|}{-} & - & - & - & .601 &	.731 &	.619 & .735 & .734 &	.623 &	.697 &	.665 \\ \midrule
 Ours, SemEval \tiny{(RoBERTa)} & \multicolumn{1}{l|}{\tiny{Zero-shot}} & \textbf{.710} & \textbf{.327} & \textbf{.282} & .585 & .727	& .614 & - & - & - & - & -\\
 Ours, ISEAR \tiny{(RoBERTa)} & \multicolumn{1}{l|}{\tiny{Zero-shot}}& .595 & .027 & .218 & - & - & - &	.735 &	.735 & - & - & - \\
 Ours, GE \tiny{(RoBERTa)} & \multicolumn{1}{l|}{\tiny{Zero-shot}}& .602 & .308 & .271 & - & - & - & - & - & .604 & .690 &	.660 \\
 \midrule
  Ours, EB$\leftarrow$SemEval \tiny{(RoBERTa)} & \multicolumn{1}{l|}{\tiny{Supervised}} & .834 & .594 & \textbf{.517} & - & - & - & - & - & - & - & - \\
  Ours, EB$\leftarrow$ISEAR \tiny{(RoBERTa)} & \multicolumn{1}{l|}{\tiny{Supervised}} & .836 & \textbf{.601} & .512 & - & - & - & - & - & - & - & - \\
  Ours, EB$\leftarrow$GE \tiny{(RoBERTa)} & \multicolumn{1}{l|}{\tiny{Supervised}} & \textbf{.842} & .594 & .512 & - & - & - & - & - & - & - & - \\
 \bottomrule
 \end{tabular}
 \caption{Validation Set Performance of VAD score prediction and categorical emotion class prediction.}
 \label{table:val_results}
 \vspace{-4mm}
 \end{table*}

\subsection{Hyperparameter Searching}
We follows default setting of models except maximum sequence length of Ours, (EB$\leftarrow$SemEval, ISEAR, GoEmotions with RoBERTa-Large).\\
The default settings are as follows: learning rate learning rate 3e-05, maximum sequence length 256, total update 10000, update frequency 4, warmup proportion 0.1, BertAdam for optimizer, and dropout 0.1. 
For supervised setting, the learning rate for 5 epochs when freezing the parameters is 3e-03 and learning rate after freezing is 5e-06. The warmup proportion during this process is 0.001. For fine-tuning experiment with 5\% of training set size, warmup proportion for RoBERTa baseline is 0.01 and number of epochs for freezing is 10 for ISEAR in the purpose of stable fine-tuning process and faster convergence.

\subsection{Dataset Details} 
In our experiment, we use four types of emotion datasets: Emobank ~\footnote{\scriptsize{https://github.com/JULIELab/EmoBank}}, SemEval: ~\footnote{\scriptsize{https://competitions.codalab.org/competitions/17751\#learn\_the\_details-datasets}},  ISEAR~\footnote{\scriptsize{http://www.affective-sciences.org/index.php/download\_file/view/395/296/}}, and GoEmotions~\footnote{\scriptsize{https://github.com/google-research/google-research/tree/master/goemotions}}. 
We include all the original datasets and data splitting is done as follows.
We use the train, validation, test split of EmoBank, SemEval and GoEmotions published from the authors.
In case of ISEAR, we split 7:1.5:1.5 with random seed 42 using train\_test\_split function in sklearn library, in stratified fashion to retain ratio between classes.

\subsection{Experimental Details}
In all experiment, we specifically use RoBERTa-Large~\footnote{\scriptsize{https://huggingface.co/transformers/model\_doc/roberta.html}} and BERT-Large trained on cased English text using Whole-Word-Masking~\footnote{\scriptsize{https://huggingface.co/transformers/model\_doc/bert.html}}. The details of model structure are described in model library~\footnote{\scriptsize{https://huggingface.co/transformers/pretrained\_models.html}}. RoBERTa-Large contains 355M trainable parameters and BERT-Large has 340M.
The batch size is set to 32, we stop fine-tuning all of the layers when the validation loss and metrics are converged. 
We use 1 GPU (RTX 6000 Ti), and take less than 5 hours for each runs. All evaluation measures in test and validation split results are average of 5 runs.

\end{document}